\documentclass[conference, 9pt]{IEEEtran}
\IEEEoverridecommandlockouts

\usepackage{cite}
\usepackage{amsmath,amssymb,amsfonts}
\usepackage{algorithmic}
\usepackage{graphicx}
\usepackage{textcomp}
\usepackage{xcolor}
\def\BibTeX{{\rm B\kern-.05em{\sc i\kern-.025em b}\kern-.08em
x    T\kern-.1667em\lower.7ex\hbox{E}\kern-.125emX}}

\usepackage{subcaption}
\usepackage{multirow}
\usepackage{booktabs}
\usepackage{url}
\usepackage{hyperref}
\usepackage{makecell}
\usepackage{svg}
\usepackage{soul}
\usepackage{tikz}  %

\begin{document}

\title{PR{\sc imu}S: Pretraining IMU Encoders with\\Multimodal Self-Supervision}

\author{
\IEEEauthorblockN{
    \centering\begin{tabular}{cccc}
      Arnav M. Das\IEEEauthorrefmark{2}\textsuperscript{*}\thanks{\textsuperscript{*}Work has been done during the author’s internship at Nokia Bell Labs.} & 
      Chi Ian Tang\IEEEauthorrefmark{3} & 
      Fahim Kawsar\IEEEauthorrefmark{3}\IEEEauthorrefmark{4} &
      Mohammad Malekzadeh\IEEEauthorrefmark{3}
      \end{tabular}
}
\IEEEauthorblockA{
\centering\begin{tabular}{ccc}
\IEEEauthorrefmark{3}Nokia Bell Labs Cambridge, UK & \IEEEauthorrefmark{2}University of Washington, USA &
\IEEEauthorrefmark{4}University of Glasgow, UK 
\end{tabular}
}
\IEEEauthorblockA{
\centering\begin{tabular}{cc}
arnavmd2@uw.edu & \{ian.tang, fahim.kawsar, mohammad.malekzadeh\}@nokia-bell-labs.com
\end{tabular}
}
}

\maketitle

\begin{abstract}
Sensing human motions through Inertial Measurement Units (IMUs) embedded in personal devices has enabled significant applications in health and wellness. Labeled IMU data is scarce, however, unlabeled or weakly labeled IMU data can be used to model human motions. For video or text modalities, the ``pretrain and adapt'' approach utilizes large volumes of unlabeled or weakly labeled data to build a strong feature extractor, followed by adaptation to specific tasks using limited labeled data. However, pretraining methods are poorly understood for IMU data, and pipelines are rarely evaluated on out-of-domain tasks. We propose PRIMUS: a method for PRetraining IMU encoderS that uses a novel pretraining objective that is empirically validated based on downstream performance on both in-domain and out-of-domain datasets. The PRIMUS objective effectively enhances downstream performance by combining self-supervision, multimodal, and nearest-neighbor supervision. With fewer than 500 labeled samples per class, PRIMUS  improves test accuracy by up to 15\%, compared to state-of-the-art baselines.
To benefit the broader community, we have open-sourced our code at \textcolor{blue}{\href{https://github.com/nokia-bell-labs/pretrained-imu-encoders}{github.com/nokia-bell-labs/pretrained-imu-encoders}}.\\
\end{abstract}


\section{Introduction}
Wearable devices embed Inertial Measurement Unit (IMU) sensors, including accelerometers and gyroscopes, which track the movement, acceleration, and orientation of the human body. When modeled using machine learning~(ML) methods, IMU data provides valuable insights into human physical and emotional behaviors, playing a crucial role in health monitoring and overall well-being~\cite{liang2014energy, ghayvat2015wellness, rachuri2010emotionsense, zhang2022coughtrigger, pacheco2023towards}. For example, step-counting data from IMU sensors is among the most effective indicators of cognitive impairment progression in elderly individuals~\cite{chen2019developing}. This potential has motivated the community to collect vast amounts of IMU data in time-series form. However, obtaining large amounts of {\em labeled} IMU data remains a major challenge, as IMU time series are inherently difficult to interpret and annotate, even for experts~\cite{yuan2024self}.

A promising solution for label scarcity is the ``pretrain once, adapt many times'' approach. This involves initially training an {\em encoder} on a large corpus of unlabeled or weakly labeled data. Afterward, a smaller ML model is trained on top of the (typically frozen) encoder for specific tasks, using relatively small amounts of labeled data. While this approach has shown significant success in image, video, audio, and natural language processing, its potential in the context of IMU data remains underexplored, primarily because of the challenges associated with curating large volumes of high-quality datasets.

\begin{figure}
    \includegraphics[width=0.49\textwidth]{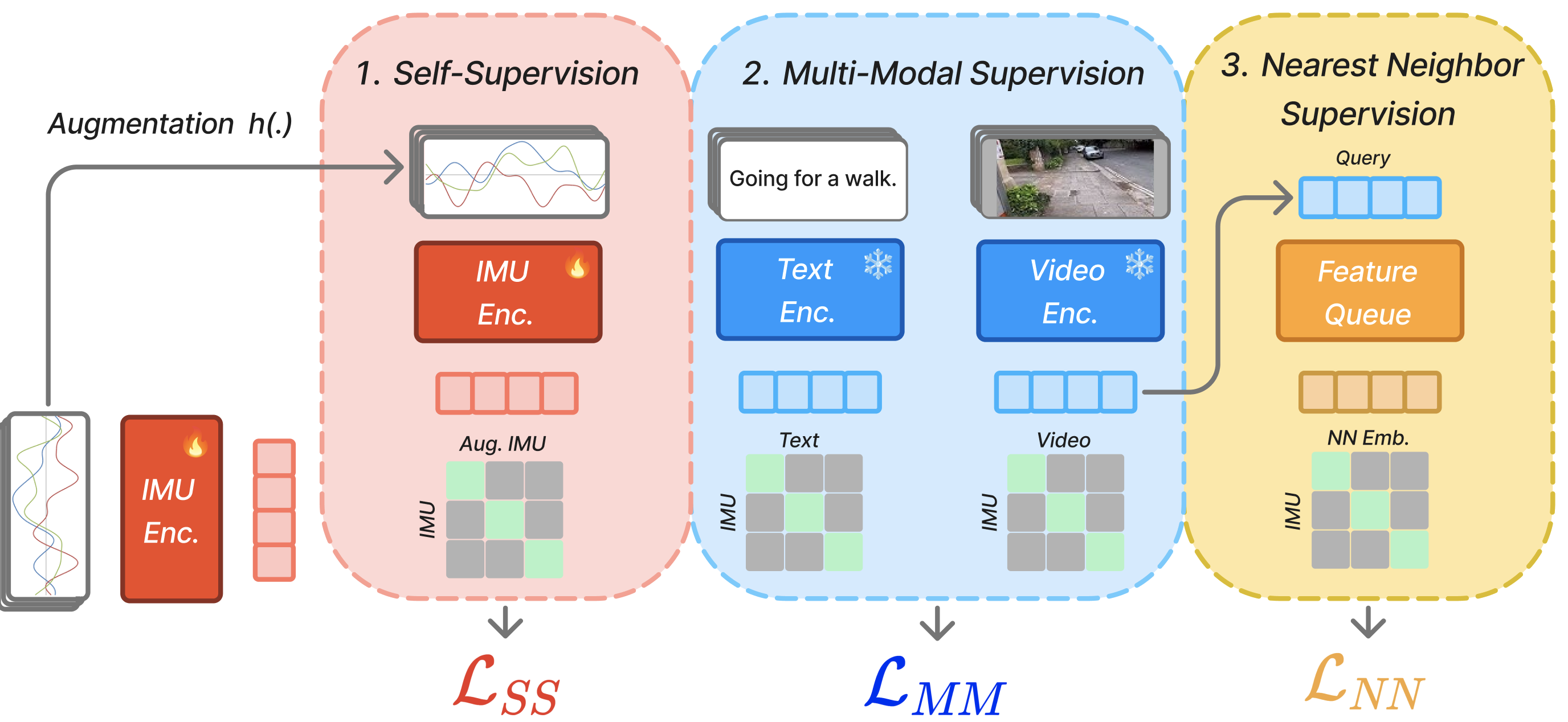}
    \caption{\small \textbf{PRIMUS Overview.} We use a multi-objective pretraining including three terms, $\mathcal{L}_{SS}, \mathcal{L}_{MM},$ and $\mathcal{L}_{NN}$. Self-supervised losses encourage the IMU encoder to be augmentation invariant, while multimodal and nearest neighbor losses align the IMU data to co-occurring video and/or text data. We use open-source models for the text and video encoders.}
    \label{fig:main_objective}
    \vspace{-.15in}
\end{figure}

Difficulties in collecting labeled data motivate {\em representation learning} methods for IMU encoders by using supervisory signals from IMU data itself ({\em self-supervised learning}), or other concurrent modalities ({\em multimodal learning}). 
Self-supervised~(SS) learning approaches based on multi-task learning~\cite{saeed2019multi, tang2021selfhar}, contrastive learning~\cite{tang2021exploringcontrastivelearninghuman, xu2023augmentation}, and masked reconstruction~\cite{haresamudram2020masked}, remain unevaluated in cross-domain use cases. Multimodal~(MM) learning has become popular in the field of representation learning~\cite{radford21a, li2022blip, wav2clip, Luo2021CLIP4Clip, elizalde2022clap}, and has recently been used for pretraining IMU encoders by utilizing supervisory signals from multiple devices~\cite{jain2022collossl, song2024learning} or multimodal data~\cite{deldari2024crossl}.  IMU2CLIP~\cite{imu2clip} aligns the latent representations of IMU data with those coming from text annotations or those from egocentric videos, where they show enhanced capabilities in multimodal data retrieval.

While both classes of representation learning approaches, i.e., SS and MM, have shown promising results, neither fully leverages the diverse sources of information present in IMU time series. Given the successful application of the synergistic relationship between self-supervised and multimodal learning in the fields of computer vision and natural language processing~\cite{mu2021slip, li2022supervision, verma2023effective}, and with the recent public release of the large multimodal dataset EgoExo4D~\cite{egoexo4d}, which includes synchronized video, text, and IMU segments, we aim to explore this combination for pretraining IMU encoders.  

In this paper, we propose \textbf{PRIMUS} (see Figure~\ref{fig:main_objective}): a novel method for \textbf{PRetraining IMU encoderS} that produces transferable representations. PRIMUS employs a multi-objective representation learning strategy that combines SS and MM losses to pretrain an IMU encoder. Pretrained on the recently released EgoExo4D dataset~\cite{egoexo4d}, we assess the effectiveness of our strategy by evaluating how well the PRIMUS IMU encoder performs on both {\em in-domain} and {\em out-of-domain} classification tasks with only a small amount of labeled data (i.e., few-shot learning). A consistent performance improvement of up to 15\% in test accuracy, compared to state-of-the-art multimodal and self-supervised training methods, was observed throughout different levels of data availability. Our ablation studies also showcase the superior performance of our combined training objective. Our results demonstrate that PRIMUS enables encoders to learn highly transferable representations, allowing for various future adaptations.

\section{Methodology}
Let $\mathcal{I}$ denote an {\em encoder} that takes a segment of multivariate IMU time series 
as input and generates a latent representation 
as output. As shown in Fig.~\ref{fig:main_objective}, we train $\mathcal{I}$ with three objectives: {\em self-supervision loss} ($\mathcal{L}_{SS}$), {\em multimodal loss} ($\mathcal{L}_{MM}$), and {\em nearest-neighbour loss} ($\mathcal{L}_{NN}$). (1)~$\mathcal{L}_{SS}$ ensures that $\mathcal{I}$ remains invariant to noise, similar to those that are introduced by slight changes in sensor position or type~(\S\ref{subsec_ssl}). (2)~$\mathcal{L}_{MM}$ pushes IMU representations towards aligned text and video representations, allowing $\mathcal{I}$ to learn the rich semantic information present in other modalities~(\S\ref{subsec_mms}). (3)~$\mathcal{L}_{NN}$ uses the closest examples in representation space as positive pairs, enabling the model to leverage natural data similarities for more adaptive contrastive learning~(\S\ref{subsec_nnl}).

\begin{figure}
    \centering
    \begin{minipage}{0.41\columnwidth}
    \includegraphics[width=\columnwidth]{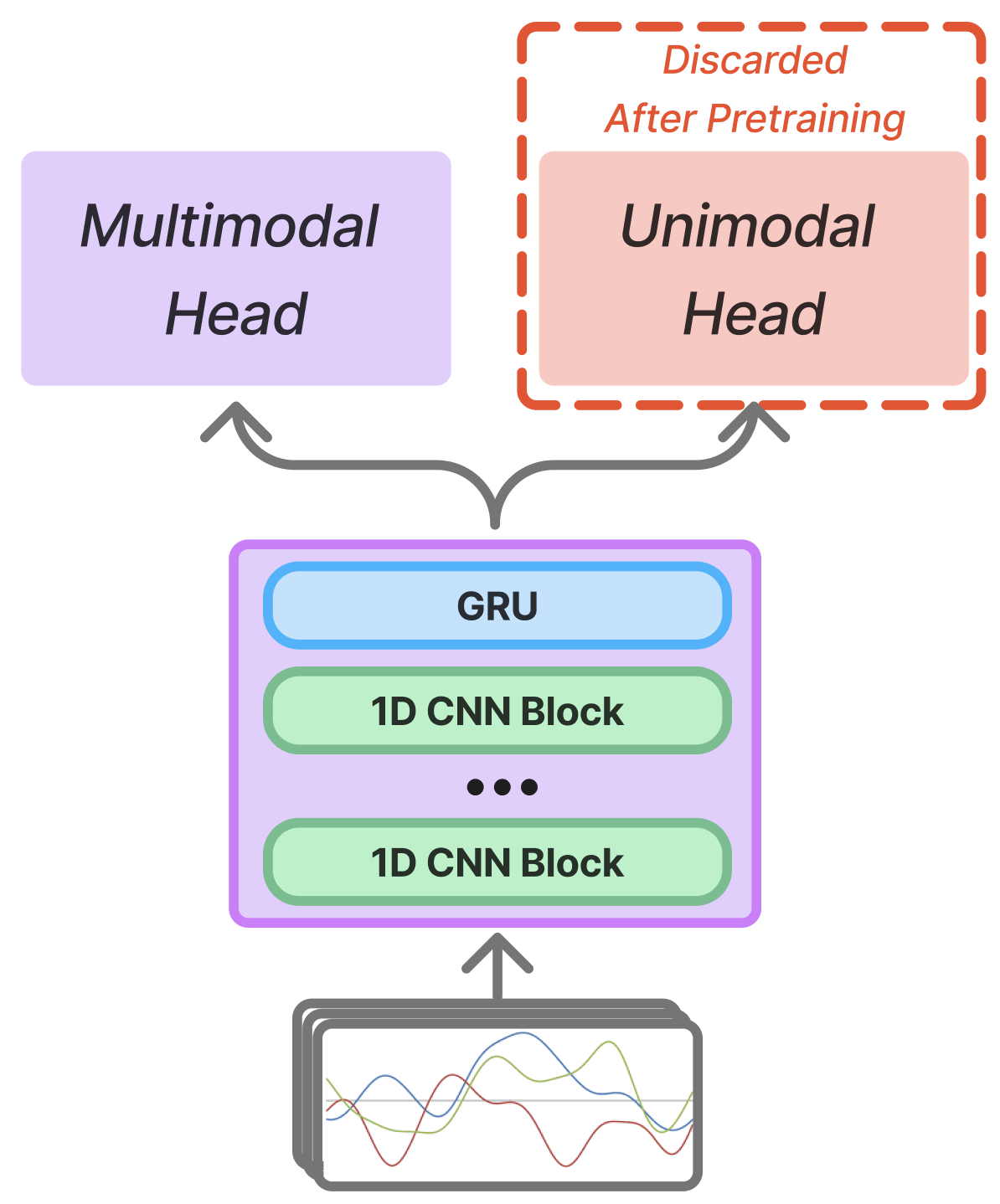}  
    \end{minipage}\hfill
    \begin{minipage}{0.5\columnwidth}
    \caption{\small \textbf{The architecture of IMU Encoder $\mathcal{I}$.} The backbone consists of both 1D-CNN and GRU layers. During pretraining, the IMU encoder has two MLP heads: one for multimodal loss and the other for unimodal loss. After pre-training, only the output of the multimodal head is kept for training downstream tasks, as it offers a more generalized latent representation. The architecture is adopted from~\cite{imu2clip}.}\label{fig_imu_enc}
    \end{minipage}
\end{figure}

In our implementation~(see Fig.~\ref{fig_imu_enc}), $\mathcal{I}$ is a Stacked RNN consisting of convolutional, group normalization, and max-pooling layers, topped with a GRU layer, based on the architecture of the IMU2CLIP model~\cite{imu2clip}, with a total of 1.4M parameters. Our main motivation for this architecture is its efficiency in deployment on mobile and wearable devices, which are the target platforms for collecting IMU data~\cite{lu2022local}. Moreover, a Stacked RNN has shown effective generalization performance in processing ML tasks on IMU data. During pre-training, $\mathcal{I}$ has two MLP heads: the first head is used to compute the {\em unimodal} self-supervision loss and the second head is used to compute the {\em multimodal} loss. For downstream tasks, only the latter is retained as it provides a richer latent representation. 

For pretraining, we use the EgoExo4D dataset~\cite{egoexo4d}, a multimodal dataset containing IMU data from head-placed sensors, egocentric videos, and free-form text annotations. After pre-processing, this dataset consists of around 250K segments, each of 5-second length, providing aligned IMU, video, and text triplets. We denote the pre-training dataset as $\mathcal{D} = \{(m_i, v_i, t_i)\}_{i=1}^N$ where $m_i$, $v_i$, $t_i$ correspond to a single segment of time-aligned IMU, video, and text, respectively.

\subsection{Self-Supervision}\label{subsec_ssl}
The self-supervised learning objective is an unimodal loss that encourages the representations of augmented versions of the same data to be similar (the first block, shown in red in Fig.~\ref{fig:main_objective}). For data augmentation, we define a stochastic transformation module $h(.)$ consisting of two transformations: (1) scaling by a random factor and (2) reversing the direction of time (see~\cite{tang2021exploringcontrastivelearninghuman} for more details). These transformations were chosen after evaluating all pairs of augmentations proposed in \cite{tang2021exploringcontrastivelearninghuman}. Given a batch $B = \{(m_i, v_i, t_i)\}_{i=1}^n$, and considering $\tau$ as a learnable temperature parameter, the self-supervised objective, adapted from SimCLR~\cite{simclr, tang2021exploringcontrastivelearninghuman}, can be formally expressed as
\begin{equation*}
    \mathcal{L}_{SS}(B) = - \frac{1}{n} \sum_{i = 1}^{n} \log \left ( \frac{\exp\big(\mathcal{I}(m_i) \cdot \mathcal{I}(h(m_i))\big)^{1/\tau}}{\sum_{k=1}^{n} \exp\big(\mathcal{I}(m_i) \cdot \mathcal{I}(h(m_k))\big) ^ {1/\tau}} \right ), 
\end{equation*}

\subsection{Multimodal Supervision}\label{subsec_mms}
We use multimodal learning (the second block, shown in blue in Fig.~\ref{fig:main_objective}) in order to allow the IMU encoder to learn semantic features that are present in rich modalities such as text and video, but difficult to learn with self-supervision alone~\cite{imu2clip}. Many open-source video and text encoders have been pretrained on web-scale data and can be used to produce rich representations for the video/text in each frame. Throughout this paper, we use an open-source video encoder $\mathcal{V}$ and text encoder $\mathcal{T}$ produced by CLIP4Clip~\cite{Luo2021CLIP4Clip} to instantiate our multimodal learning objective, since this model is designed to handle short video clips and is readily available. Given a batch $B= \{(m_i, v_i, t_i)\}_{i=1}^n$, the multimodal loss has two components which can be expressed as 
\begin{equation}
    \mathcal{L}_{m2v}(B) = -\frac{1}{n}\sum_{i = 1}^{n} \log \left ( \frac{\exp\big(\mathcal{I}(m_i) \cdot \mathcal{V}(v_i)\big)^{1/\tau}}{\sum_{j=1}^{n} \exp\big(\mathcal{I}(m_i) \cdot \mathcal{V}(v_j)\big) ^ {1/\tau}} \right ),  
\end{equation}
and
\begin{equation}
    \mathcal{L}_{m2t}(B) = - \frac{1}{n}\sum_{i = 1}^{n} \log \left ( \frac{\exp\big(\mathcal{I}(m_i) \cdot \mathcal{T}(t_i)\big)^{1/\tau}}{\sum_{j=1}^{n} \exp\big(\mathcal{I}(m_i) \cdot \mathcal{T}(t_j)\big) ^ {1/\tau}} \right ),
\end{equation}
\noindent where again $\tau$ is a learnable temperature. Intuitively, $\mathcal{L}_{m2v}$ (or $\mathcal{L}_{m2t}$) encourage $\mathcal{I}$ to map IMU data to representations that are close to corresponding video (or text) representations in the latent space. 

We use $\mathcal{L}_{MM}(B) = \mathcal{L}_{m2v}(B) + \mathcal{L}_{m2t}(B)$ as our MM objective.

\begin{figure}
    \centering
    \includegraphics[width=0.8\columnwidth]{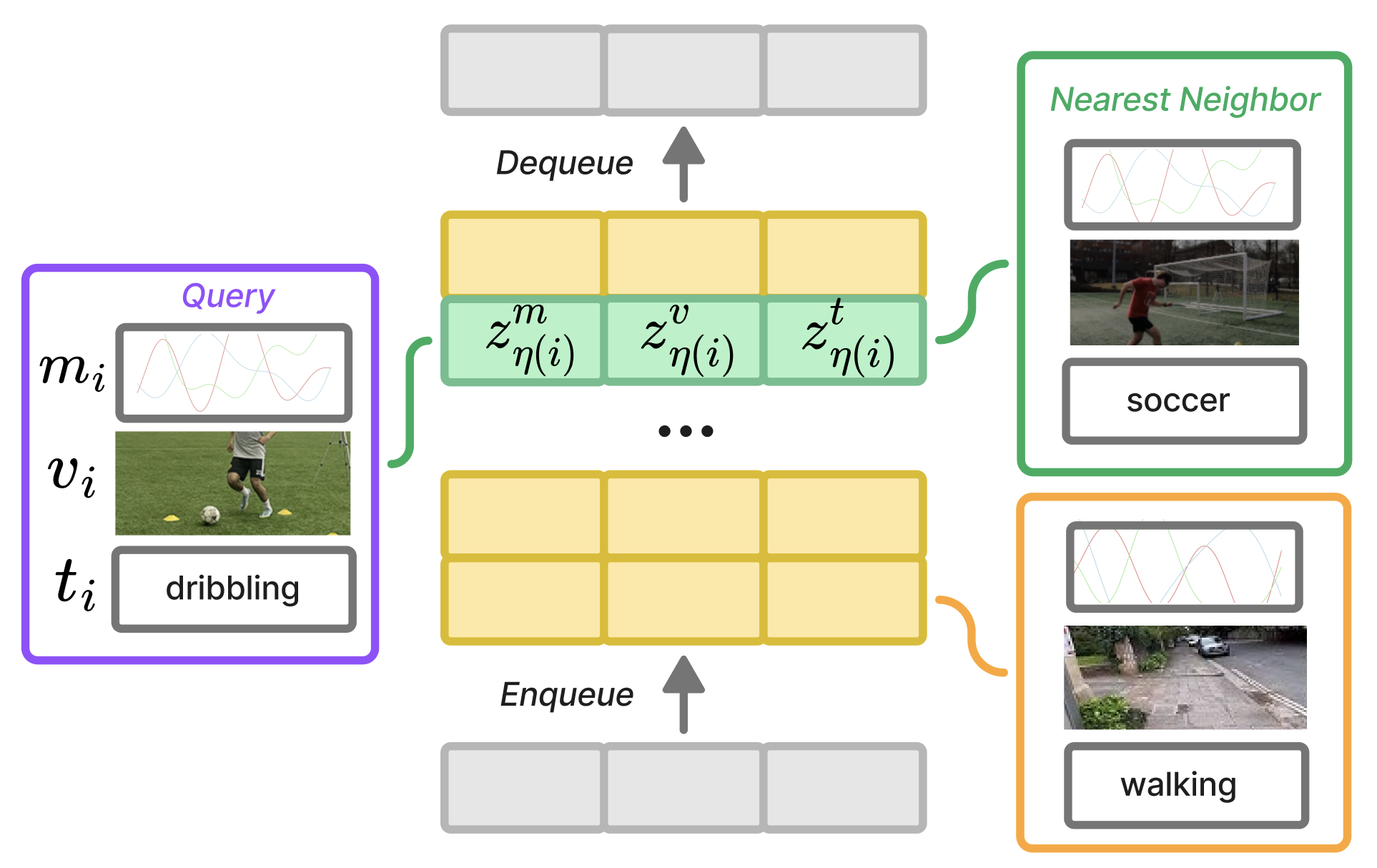}  
    \caption{\small \textbf{Nearest neighbor supervision.} Given a query segment, we retrieve the most similar segment in the queue, based on video-to-video similarity, and use all modalities to derive supervisory signals for the IMU segment. Features are retrieved from a fixed-size queue.}\label{fig:queue}
\end{figure}

\begingroup
\renewcommand{\arraystretch}{1.1}

\begin{table*}[htp]
\centering
\caption{\textbf{Downstream Tasks.} A summary of the classification datasets used for downstream evaluation. Unlike previous work on IMU representation learning, we consider tasks that have IMU data collected from unseen devices and have novel output domains.}\label{tab:datasets}
\resizebox{\textwidth}{!}{%
\begin{tabular}{l|l|l|l|l}
\toprule
\textbf{Test Set} & \textbf{Activities} & \textbf{Input Domain} & \textbf{Output Domain} & \textbf{Sample Size} \\
\midrule
EgoExo4D
~\cite{egoexo4d}  & {8}: \{play music, cook, medical test, perform CPR, repair bike, climb rock,  soccer,  dance\} & \multirow{1}{*}{Same} & \multirow{1}{*}{Same} & Train: 195K--Test: 53K \\
Ego4D~\cite{ego4d}    & {10}: \{play music, cook, eat, clean, carpenter, craft, farmer, household, walk, construction\} &  \multirow{1}{*}{Same} & \multirow{1}{*}{\em Different} & Train: 555K--Test: 57K\\
REALWORLD~\cite{sztyler2016body} & {8}: \{climbing up, climbing down, jumping, lying down, run, walk, sit, down\} & \multirow{1}{*}{\em Different} & \multirow{1}{*}{\em Different} & Train: 8.3K--Test: 2.6K \\
\bottomrule
\end{tabular}%
}

\end{table*}
\endgroup

\begin{figure*}[htp!]
\centering
    \begin{subfigure}{.32\textwidth}
        \includegraphics[width=\linewidth]{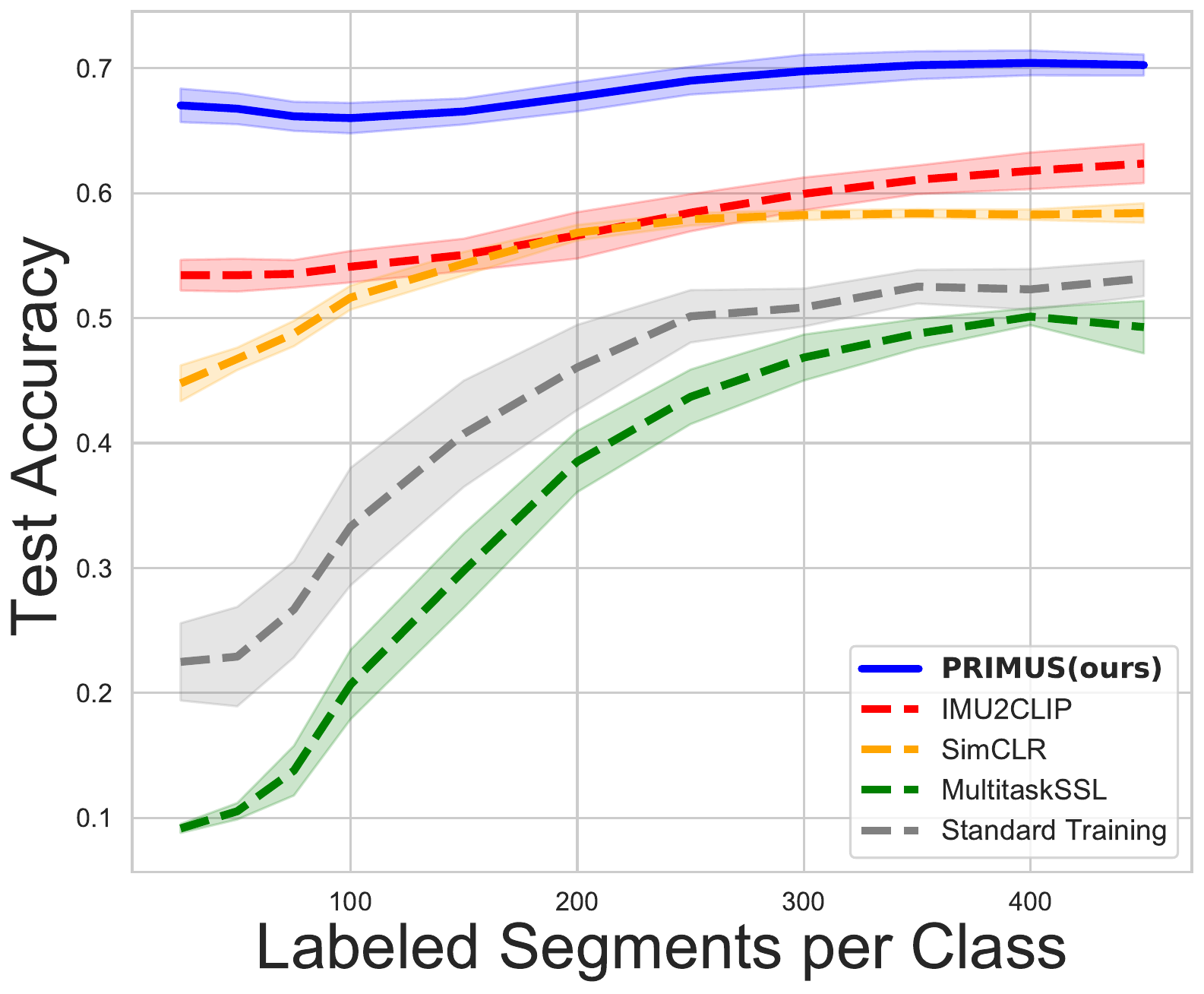}
        \caption{EgoExo4D Results}
    \end{subfigure}\quad
    \begin{subfigure}{.32\textwidth}
        \includegraphics[width=\linewidth]{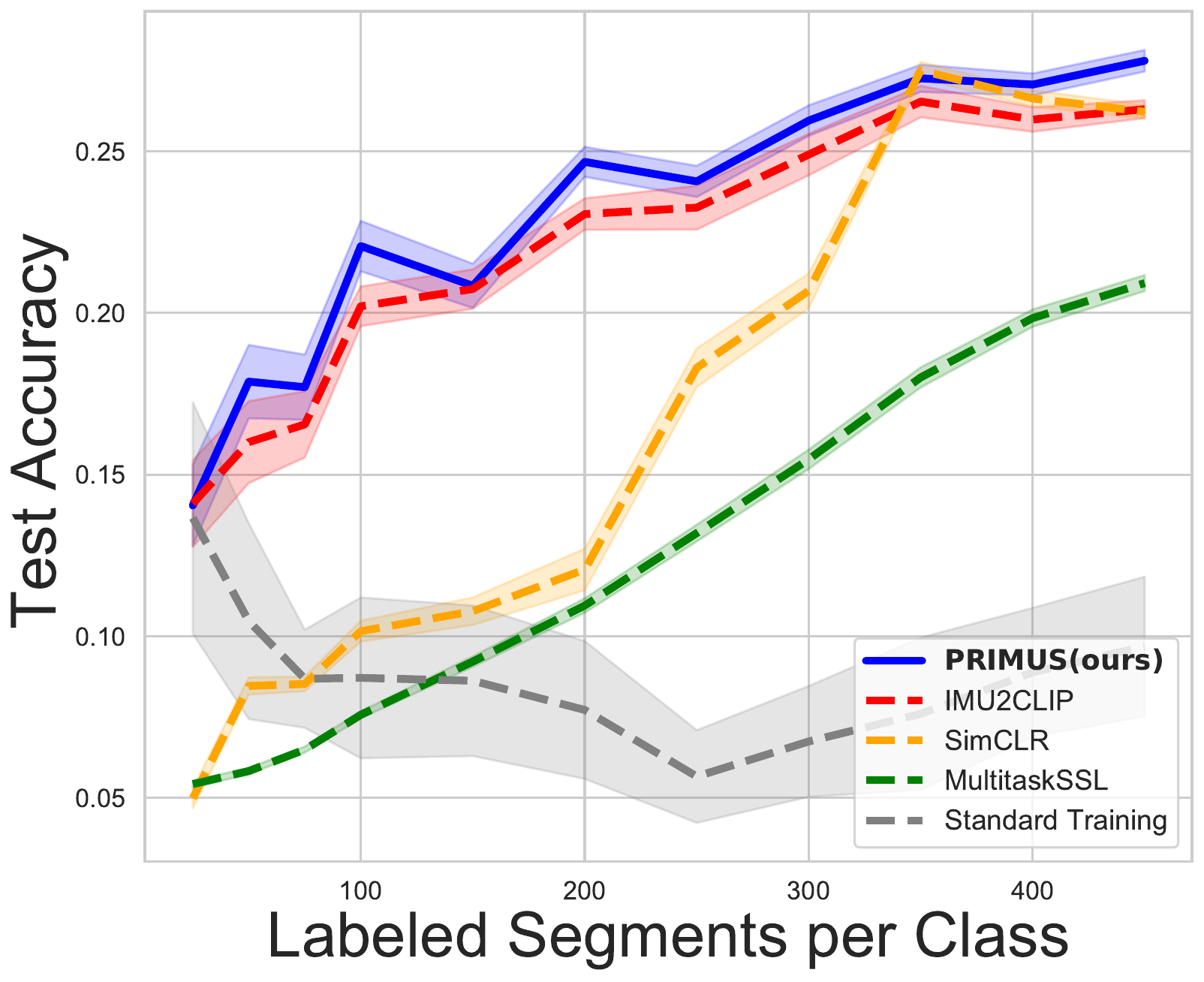}
        \caption{Ego4D Results}
    \end{subfigure}\quad
    \begin{subfigure}{.32\textwidth}
        \includegraphics[width=\linewidth]{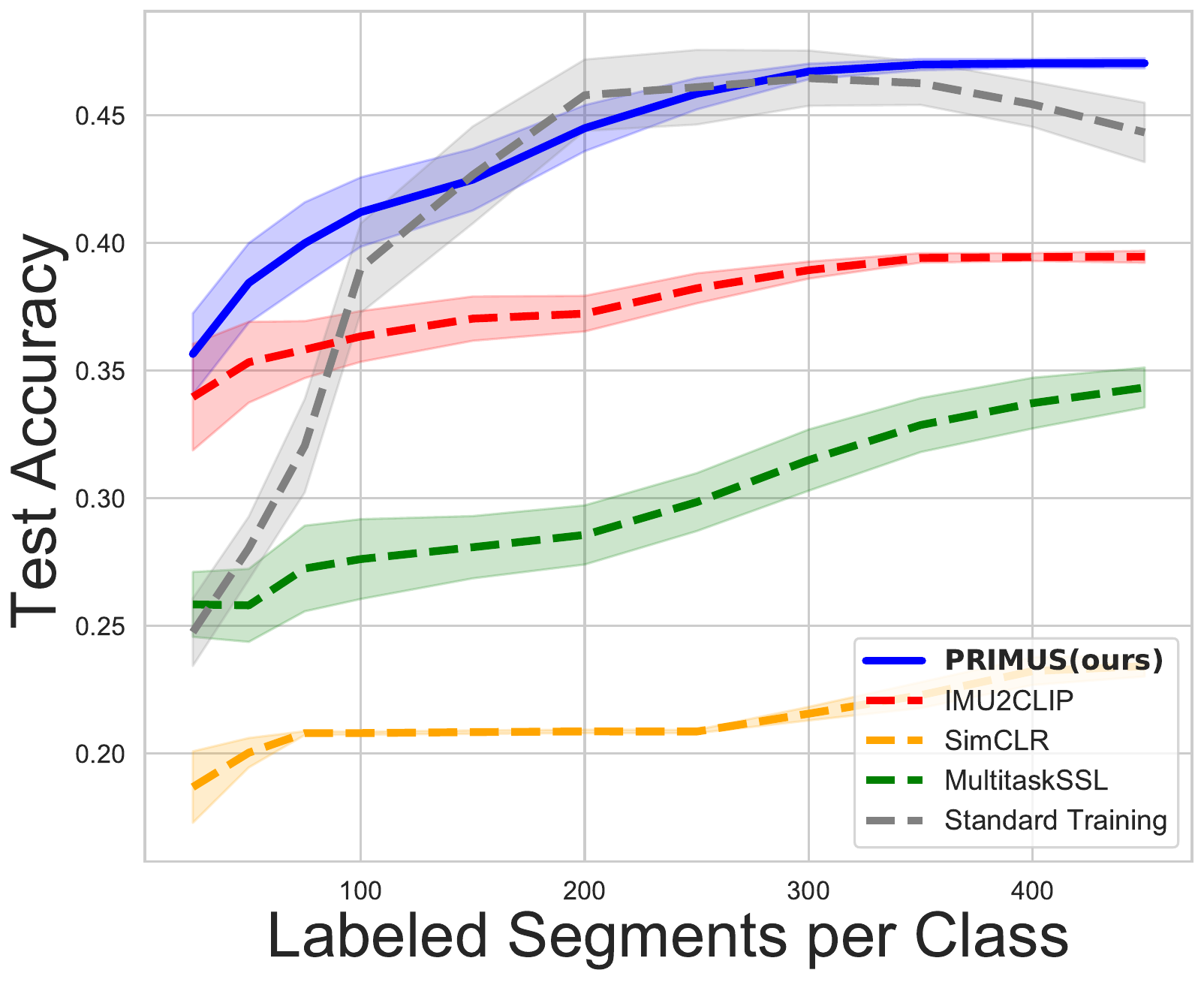}
        \caption{REALWORLD Results}
    \end{subfigure}
    \caption{{\small \textbf{Main Results.} We report the few-shot learning performance of pretrained models on various classification datasets. PRIMUS generally outperforms self-supervised methods (SimCLR, MultitaskSSL), and prior multimodal methods (IMU2CLIP), as well as training a randomly initialized model (standard training). The standard error is computed over 5 trials.}}
    \label{fig: proxy}
\end{figure*}

\subsection{Nearest Neighbor Supervision} \label{subsec_nnl}

The loss terms introduced so far, $\mathcal{L}_{SS}$ and $\mathcal{L}_{MM}$, both derive supervision from within the same triplet segment. To increase the diversity of supervision and go beyond a single instance, we leverage nearest-neighbor supervision~\cite{li2022supervision, nnclr} (shown in the rightmost block in orange in Fig.~\ref{fig:main_objective} and in detail in Fig.~\ref{fig:queue}). During training, we maintain a feature queue $\mathcal{Q} = \{(z_j^{m},z_j^{v},z_j^{t})\}_{j=1}^K$, where $z_j^m$, $z_j^v$, and $z_j^t$ are cached representations of IMU, video, and text produced from their respective encoders. For every given instance $(m_i, v_i, t_i)$ in a batch $B$, we define
\begin{equation}
\eta(i) = \text{argmax}_{k \in [K]} \left(z_k^{v} \cdot \mathcal{V}(v_i)\right),
\end{equation}
which identifies the index $k$ in $\mathcal{Q}$ corresponding to the video embedding that is the most similar to $v_i$. We leverage the video representations for identifying the closest pairs because the video encoder is pretrained on a large dataset, and therefore produces stable representations. Also, videos capture much finer details about human activities compared to text descriptions. We illustrate the queuing mechanism for nearest neighbor retrieval in Fig.~\ref{fig:queue}. We then push $\mathcal{I}(m_i)$ close to $z_{\eta(i)}^m, z_{\eta(i)}^v, z_{\eta(i)}^t$ by  $\mathcal{L}_{NN}$, which consists of a unimodal and multimodal loss similar to $\mathcal{L}_{SS}$ and $\mathcal{L}_{MM}$ as 

\begin{equation}
\mathcal{L}_{NN}(B)
= -\frac{1}{n}\!
  \sum_{{\text{mod} \in \{m,v,t\}}}
  \sum_{i=1}^{n}
  \log\frac{
    \exp\!\big(\mathcal{I}(m_i)\cdot z_{\eta(i)}^{\text{mod}}\big)^{1/\tau}
  }{
    \sum_{j=1}^{n}
    \exp\!\big(\mathcal{I}(m_i)\cdot z_{\eta(j)}^{\text{mod}}\big)^{1/\tau}
  }.
\end{equation}

The final \textbf{multi-objective loss} that we use in PRIMUS is 
\begin{equation}
\mathcal{L}(B) = \alpha \mathcal{L}_{SS}(B) + \beta \mathcal{L}_{MM}(B) + \gamma \mathcal{L}_{NN}(B).
\end{equation}
\noindent In our experiments we set $\alpha$=$\beta$=$\gamma$=$1$, leaving the fine-tuning of hyperparameters to future studies. Thus, the results in the following sections represent lower bounds on the performance achievable with a thorough hyperparameter search.

\begin{figure*}[htp!]
\centering
    \begin{subfigure}{.32\textwidth}
        \includegraphics[width=\linewidth]{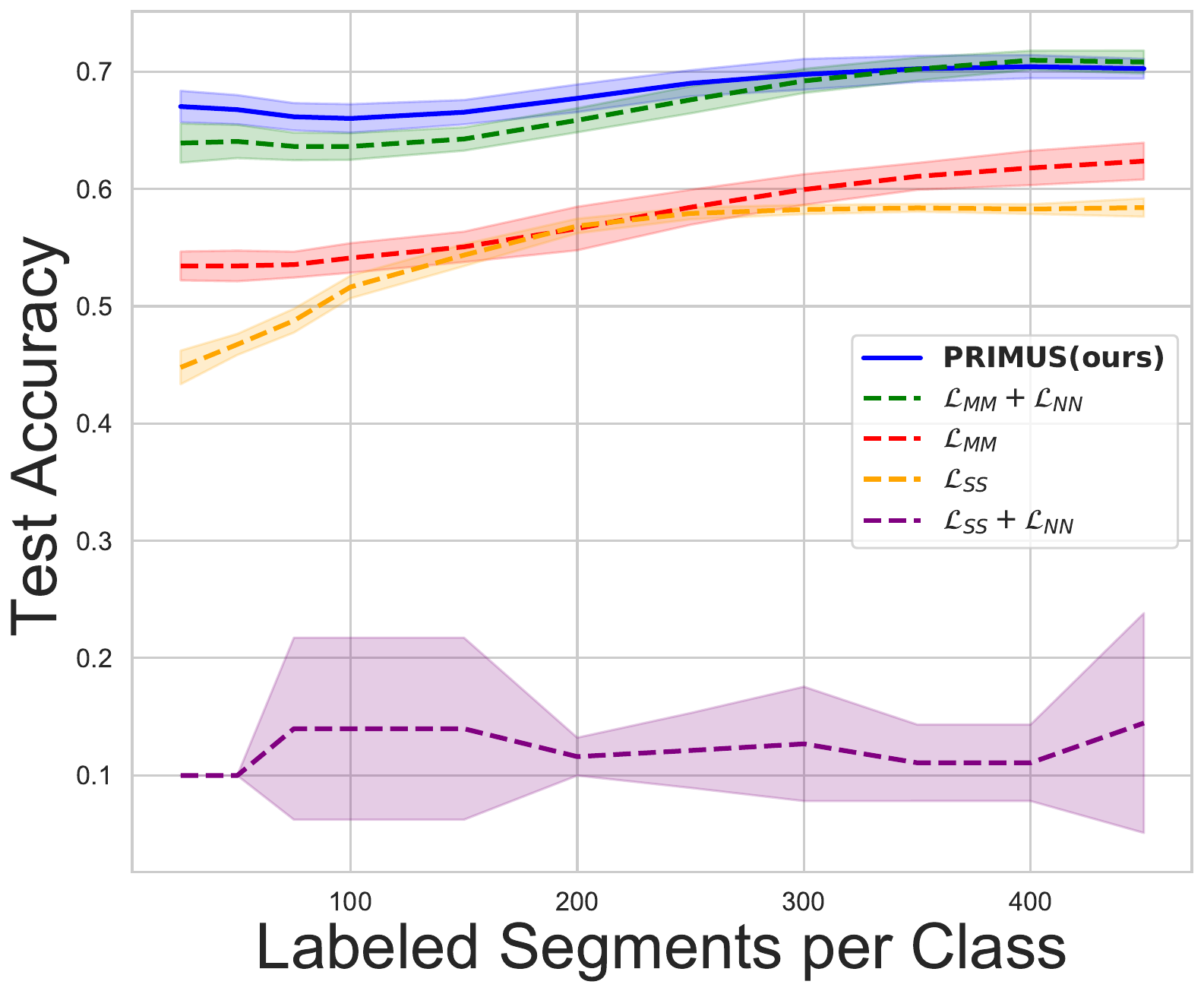}
        \caption{EgoExo4D Results}
    \end{subfigure}\quad
    \begin{subfigure}{.32\textwidth}
        \includegraphics[width=\linewidth]{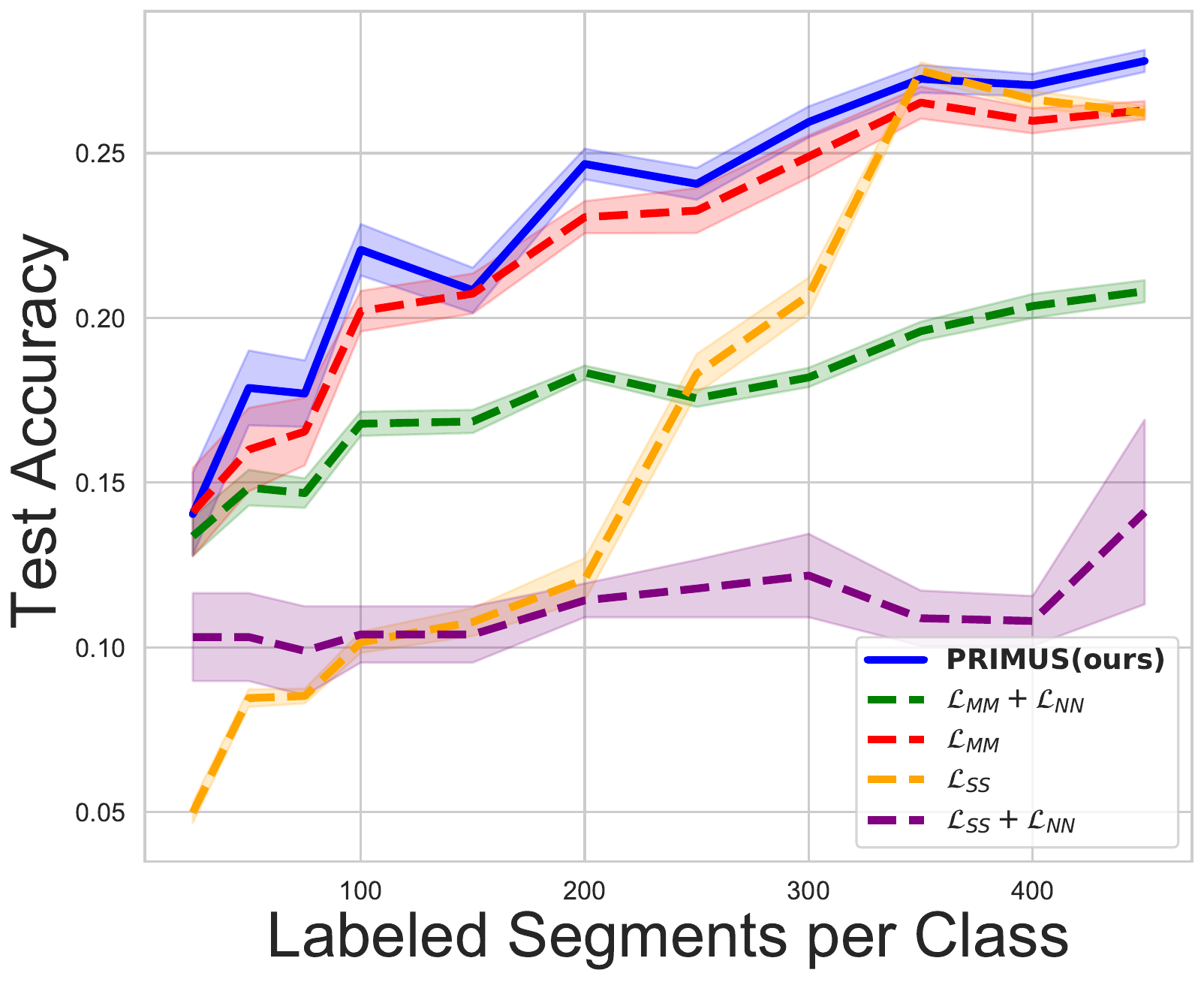}
        \caption{Ego4D Results}
    \end{subfigure}\quad
    \begin{subfigure}{.32\textwidth}
        \includegraphics[width=\linewidth]{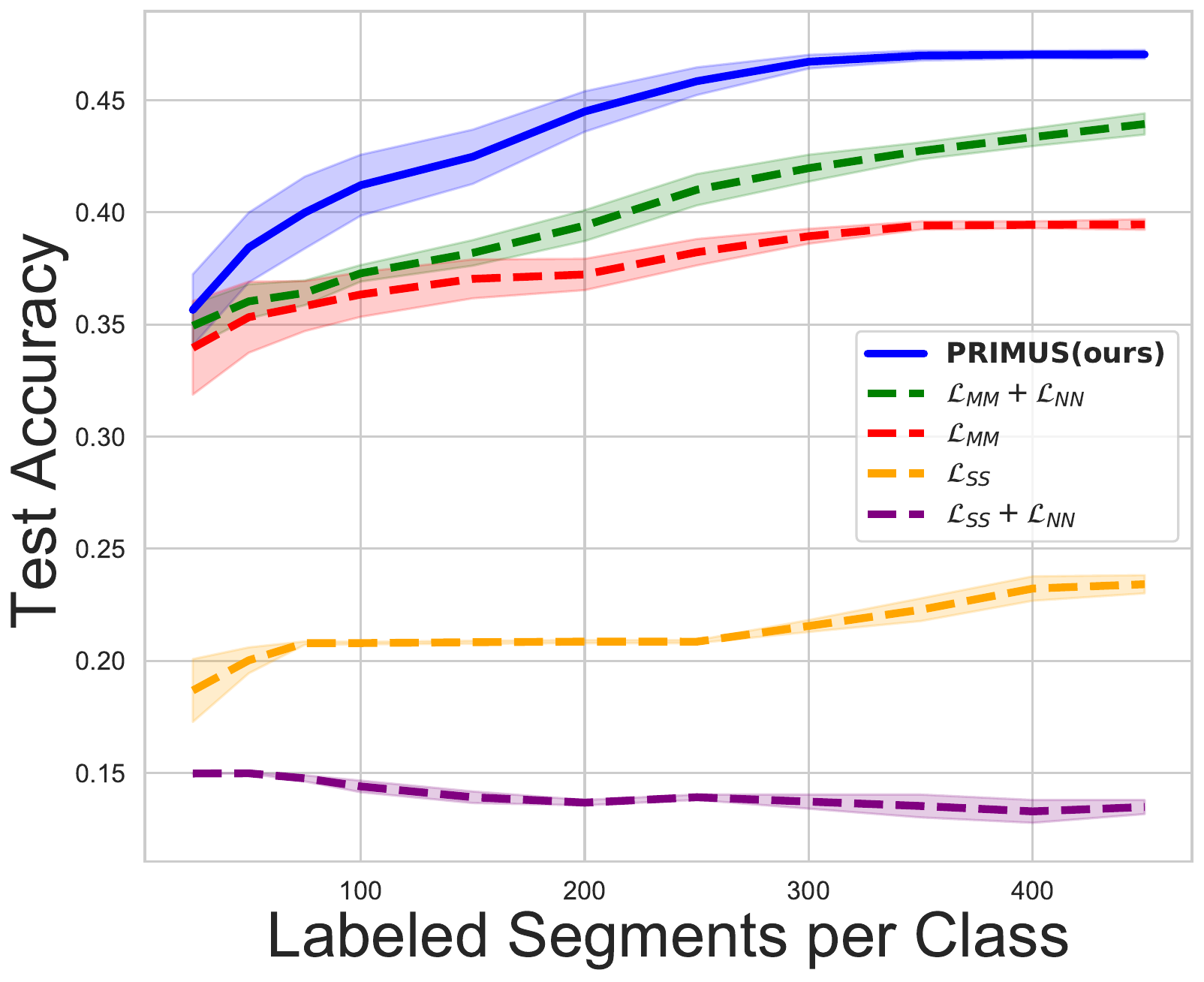}
        \caption{REALWORLD Results}
    \end{subfigure}
    \caption{{\small \textbf{Ablations.} We assess the importance of each individual term in the PRIMUS objective, by pretraining encoders with different losses and evaluating them based on few-shot learning performance. The standard error is computed over 5 trials.}}
    \label{fig:abl}
    \vspace{-0.7em}
\end{figure*}
\section{Experimental Evaluation}

For the downstream evaluations, we focus on human activity recognition tasks \textbf{using only IMU data}. We consider different levels of data scarcity by varying the number of labeled segments per class (i.e., few-shot learning). We evaluate the effectiveness of PRIMUS for a downstream task, over other pretraining baselines,  by analyzing the performance of a linear classifier on the representations produced by the IMU encoder (i.e., linear probing~\cite{radford21a}), a technique which requires few computational resources to train and retains the robustness of the pretrained encoder. 

Overall, we compare PRIMUS against other pretraining baselines~(\S\ref{sub_sec_main_res}), conduct ablations on each loss term of PRIMUS~(\S\ref{sub_sec_abl}), and evaluate data efficiency of PRIMUS~(\S\ref{sub_sec_data_ef}).

\subsection{Datasets and Setup}
All the baselines and our IMU encoder are pretrained on EgoExo4D, which contains IMU data (triaxial accelerometer and triaxial gyroscope) collected from head-placed sensors. Thus, we focus on downstream tasks that use IMU data of head-placed sensors. A summary of datasets is given in Table~\ref{tab:datasets}.

\textbf{EgoExo4D~\cite{egoexo4d}} and \textbf{Ego4D~\cite{ego4d}.} From each dataset, we choose a held-out test set for human activity recognition, where IMU data is labeled according to the activities indicated in the filenames. Note that Ego4D is captured using the same device, Project-Aria smartglass~\footnote{\url{https://www.projectaria.com}}, the same as EgoExo4D (pre-training dataset), but Ego4D includes some activities that are not present in EgoExo4D. 

\textbf{REALWORLD~\cite{sztyler2016body}.} The REALWORLD dataset is a human activity recognition dataset with 8 predefined classes, that contain data captured by various Samsung Galaxy-S4 and LG G-Watch-R placed at different positions on the body. For our analysis, we use the data from the head-placed sensor. We adopted the well-established user-based dataset-splitting strategy for our evaluations, in which data from a held-out set of users are reserved for testing, measuring the performance of the model on unseen users. This dataset also evaluates the out-of-domain performance of our pretrained models since both the set of activities and device type are different from EgoExo4D. 

\subsection{Main Results} \label{sub_sec_main_res}
We compare our \textbf{PRIMUS} against closely related training baselines. (I)~\textbf{SimCLR}~\cite{tang2021exploringcontrastivelearninghuman} is a self-supervised training method based on data augmentations. (II)~\textbf{IMU2CLIP}~\cite{imu2clip} is a multimodal training method for IMU data, corresponding to using only $\mathcal{L}_{SS}$ or $\mathcal{L}_{MM}$ as the pretraining objective on EgoExo4D. Moreover, our work leverages supervisory signals from different learning setups to train a better-performing feature extractor. Thus, we also compare PRIMUS against (III)~\textbf{MultitaskSSL}~\cite{saeed2019multi}, a well-established self-supervised approach for IMU signals. Finally, we compare PRIMUS against (IV)~\textbf{Standard Training}, which starts from a randomly initialized model and updates {\em all the parameters} (as opposed to just the final layer) with standard supervised learning. This final baseline represents the standard procedure used to train a model in the absence of a pretrained IMU encoder. 

Fig.~\ref{fig: proxy} presents the comparison of PRIMUS with all four baselines. Across all experiments, we observe that our PRIMUS model, pretrained with the joint objective, significantly outperforms any pretraining strategy previously proposed. Our method consistently outperforms all other baselines by as much as 15\% on the EgoExo4D dataset. On Ego4D, it performs on par with IMU2CLIP but still surpasses all other baselines. Additionally, on the REALWORLD dataset, our method generally outperforms all baselines, particularly in scenarios where labeled data is limited (fewer than 100 samples). Notably, standard training, which updates all the parameters, fails to generalize well in the low-data regime particularly for complex classification tasks (on EgoExo4D and Ego4D). 

\subsection{Ablations} \label{sub_sec_abl}

Fig.~\ref{fig:abl} presents an ablation study on the pretraining objectives to understand which components of the loss are most critical. We find that $\mathcal{L}_{MM}$ is a key component, indicating that future studies for developing IMU foundation models should incorporate aligned video, text, audio, and potentially other under-explored wearable sensors.

We also find that $\mathcal{L}_{NN}$ is generally helpful, but only when we have a reliable estimate of similarity. With $\mathcal{L}_{SS} + \mathcal{L}_{NN}$, we observe some form of collapse (with accuracy around 10-15\%) since this setting does not exploit any multimodal signals, and using the IMU representations itself to find similar segments from the queue can make training unstable.

Finally, while $\mathcal{L}_{SS}$ is not particularly helpful in EgoExo4D (evident from the fact that $\mathcal{L}_{MM} + \mathcal{L}_{NN}$ nearly matches the performance of PRIMUS), self-supervision seems to make a significant difference on {\em out-of-domain} tasks, offering up to 5\% of accuracy improvement in REALWORLD. We hypothesize that this is due to the fact that this loss term explicitly encourages the IMU encoder to be invariant to some of the types of noise that may be observed due to changing devices or positions on the body.

\begin{figure}
    \centering
    \begin{minipage}{0.55\columnwidth}
    \includegraphics[width=\columnwidth]{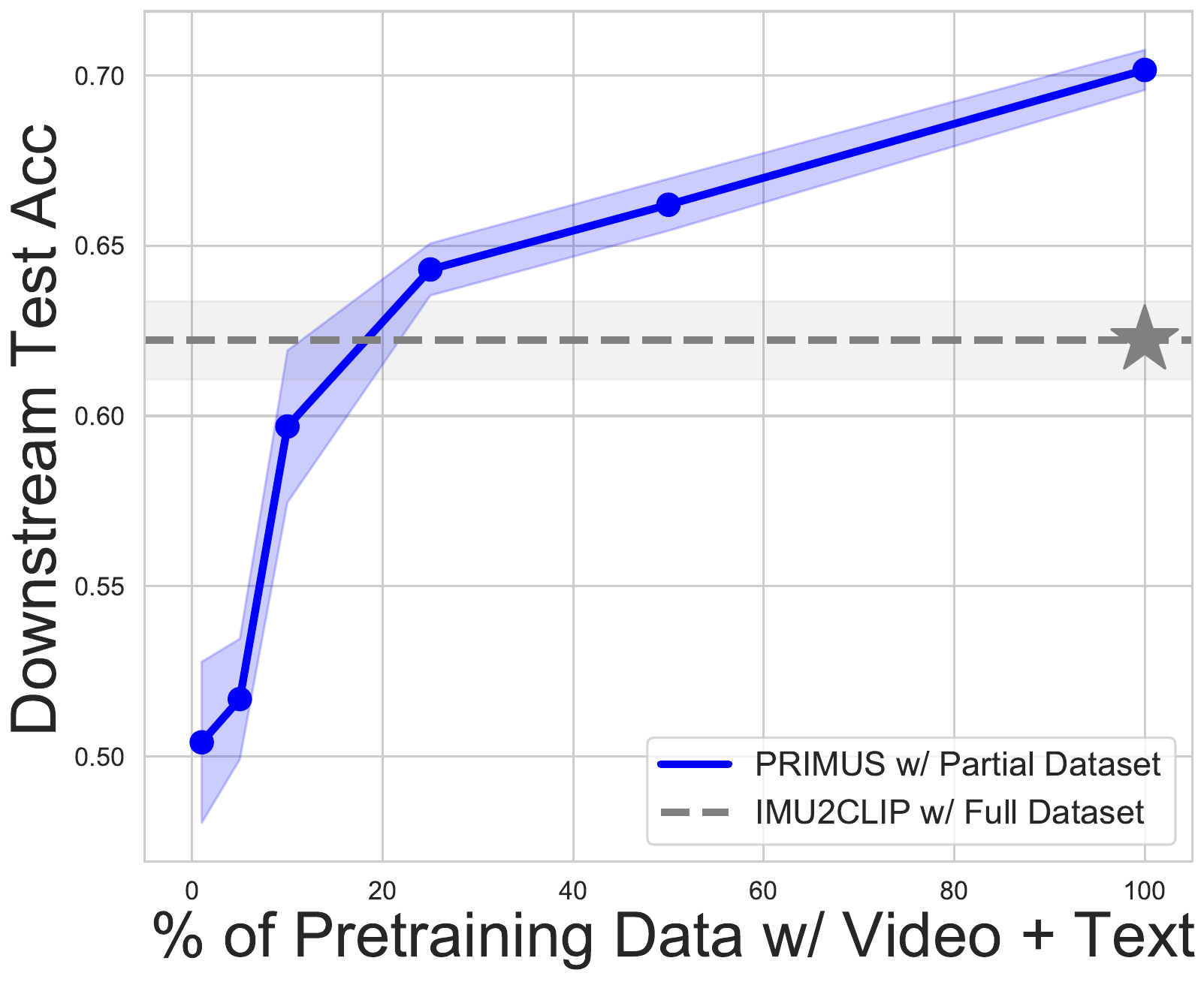}  
    \end{minipage}\hfill
    \begin{minipage}{0.4\columnwidth}
    \caption{{\small \textbf{Data Efficiency.} We report few-shot performance on the EgoExo4D classification task at 500 segments per class for PRIMUS models trained with various amounts of multimodal data. Models pretrained with the PRIMUS objective require far fewer IMU segments with aligned video/text than IMU2CLIP.}}
    \label{fig:arch}
    \end{minipage}
    \vspace{-.2in}
\end{figure}

\subsection{Pretraining Data Efficiency}\label{sub_sec_data_ef}
While we showed that the MM loss is critical for PRIMUS, obtaining large-scale IMU datasets that are temporally aligned with videos and text could be challenging. Therefore, we explore the possibility of training an effective IMU encoder using only a small portion of the data that includes aligned video and text. Specifically, we remove the aligned video and text data for different fractions of the pretraining data and evaluate the efficacy of the resulting IMU encoder in few-shot learning. Fig.~\ref{fig:arch} shows that there is no statistically significant difference between an encoder pretrained with only 10\% of the data aligned with video/text with the PRIMUS and an encoder pretrained in the style of IMU2CLIP on EgoExo4D in terms of few-shot learning performance.

\section{Conclusion and Future Work}
This paper studies pretraining objectives for building an IMU encoder that can be adapted to unseen tasks with limited labeled data. We empirically demonstrate the superiority of our pretraining method against existing approaches on in-domain and out-of-domain tasks, and identify some of the components that were critical to its success. We demonstrate that our method can be used on pretraining datasets with few samples of temporally aligned video or text.

Our work has its limitations and there are several promising directions for future work. First, while we evaluate out-of-domain downstream tasks, all of our evaluation schemes assume that the sensor position on the human body is similar to that of the pretraining set. Training a model that is capable of generalizing across human body positions is an important future direction, but pretraining datasets to enable this are not yet available. Second, our evaluation focuses on activities of medium granularity (corresponding to `\emph{actions}' according to the hierarchy of activities proposed in~\cite{moeslund2006survey}). To recognize more abstract or primitive activities, a different processing pipeline would be needed to accommodate the different time scales in which these activities occur. Further studies might be needed to adapt our proposed method for these different scenarios. 

Despite open challenges, our work contributes to developing generalizable IMU models by introducing a highly adaptable pretraining strategy. By open-sourcing our framework, we aim to encourage the community to further build upon our efforts. 
 
\clearpage

\bibliographystyle{IEEEtran}
\bibliography{strings}

\end{document}